\documentclass[letterpaper, 10 pt, conference]{ieeeconf}  

\IEEEoverridecommandlockouts                              

\usepackage{amsmath,amsfonts}
\usepackage{amssymb}

\usepackage{algorithm}
\usepackage[noend]{algorithmic}
\usepackage{color}
\usepackage{color}
\usepackage{xcolor}
\usepackage{wasysym}

\usepackage{array}
\usepackage{textcomp}
\usepackage{stfloats}

\usepackage{verbatim}
\usepackage{threeparttable}
\usepackage{graphicx}

\usepackage{cite}
\usepackage[utf8]{inputenc} 
\usepackage[T1]{fontenc}    
\usepackage{hyperref}       
\usepackage{url}            
\usepackage{booktabs}       
\usepackage{amsfonts}       
\usepackage{nicefrac}       
\usepackage{microtype}      
\usepackage{lipsum}
\usepackage{fancyhdr}       
\usepackage{graphicx}       
\usepackage{wrapfig}
\usepackage{subfigure}
\usepackage{multirow}
\usepackage{subcaption}

\newcommand{\sectionreducemargin}[1]{
\vspace{-1mm} 
\section{#1}
\vspace{-1mm} 
}

\newcommand{\subsectionreducemargin}[1]{
\vspace{-1mm} 
\subsection{#1}
\vspace{-1mm} 
}

\title{\LARGE \bf
APREBot: Active Perception System for Reflexive Evasion Robot
}

\author{Zihao Xu$^{1}$, Kuankuan Sima$^{2}$, Junhao Deng$^{3}$, Zixuan Zhuang$^{4}$, Chunzheng Wang$^{2}$, Ce Hao$^{1}$, and Jin Song Dong$^{1}$
\thanks{$^1$ School of Computing, National University of Singapore, Singapore.}
\thanks{$^2$ Department of Electrical and Computer Engineering, National University of Singapore, Singapore.}
\thanks{$^3$ School of Automation, Beijing Institute of Technology, China.}
\thanks{$^4$ School of Computer Science and Engineering, Sun Yat-Sen University, China.}
}

\begin{document}

\maketitle
\thispagestyle{empty}
\pagestyle{empty}


\begin{abstract}
Reliable onboard perception is critical for quadruped robots navigating dynamic environments, where obstacles can emerge from any direction under strict‌ reaction-time constraints. Single-sensor systems face inherent limitations: LiDAR provides omnidirectional coverage but lacks rich texture information, while cameras capture high-resolution detail but suffer from restricted field of view. We introduce APREBot (Active Perception System for Reflexive Evasion Robot), a novel framework that integrates reflexive evasion with active hierarchical perception. APREBot strategically combines LiDAR-based omnidirectional scanning with camera-based active focusing, achieving comprehensive environmental awareness essential for agile obstacle avoidance in quadruped robots. We validate APREBot through extensive sim-to-real experiments on a quadruped platform, evaluating diverse obstacle types, trajectories, and approach directions. Our results demonstrate substantial improvements over state-of-the-art baselines in both safety metrics and operational efficiency, highlighting APREBot's potential for dependable autonomy in safety-critical scenarios. Videos are available at \href{https://sites.google.com/view/aprebot/}{sites.google.com/view/aprebot/}.

\end{abstract}

\sectionreducemargin{Introduction} \label{Sec: intro}

Legged robots are increasingly expected to operate in unstructured and dynamic environments, where reliable independent onboard perception is essential for autonomous navigation and interaction. Typical sensing modalities include LiDAR~\cite{wang2025omni,xu2021fast,xu2022fast,wu2024moving}, which provides accurate range information over a wide field of view; RGB-D cameras~\cite{he2024agile,roth2024viplanner}, which capture both appearance and depth within a restricted perspective; and event cameras~\cite{zhudynamic}, which deliver high temporal resolution suited to rapid changes in the scene. These onboard sensors support core perception tasks such as environment mapping, obstacle detection, and navigation.

However, perceiving dynamic obstacles remains particularly challenging for legged robots. Such obstacles can approach from arbitrary directions, including lateral or rear blind spots, which require continuous omnidirectional awareness. At the same time, high-speed motion imposes strict reaction-time constraints, leaving only a very limited time window to detect and respond before a potential collision. Given these constraints, LiDAR provides global coverage but it cannot reliably segment objects and lacks semantic perception~\cite{alaba2022survey}, cameras offer high-resolution details but their limited field of view prevents global perception~\cite{liu2023real}, and event cameras achieve fine temporal resolution but are sensitive to noise and lack robustness~\cite{liu2023motion}. Therefore, no single sensor modality is sufficient, and combining them in a complementary manner is essential to achieve robust performance~\cite{zhou2022rgb}.

\begin{figure}[t]
    \centering
    \includegraphics[width=0.97\columnwidth]{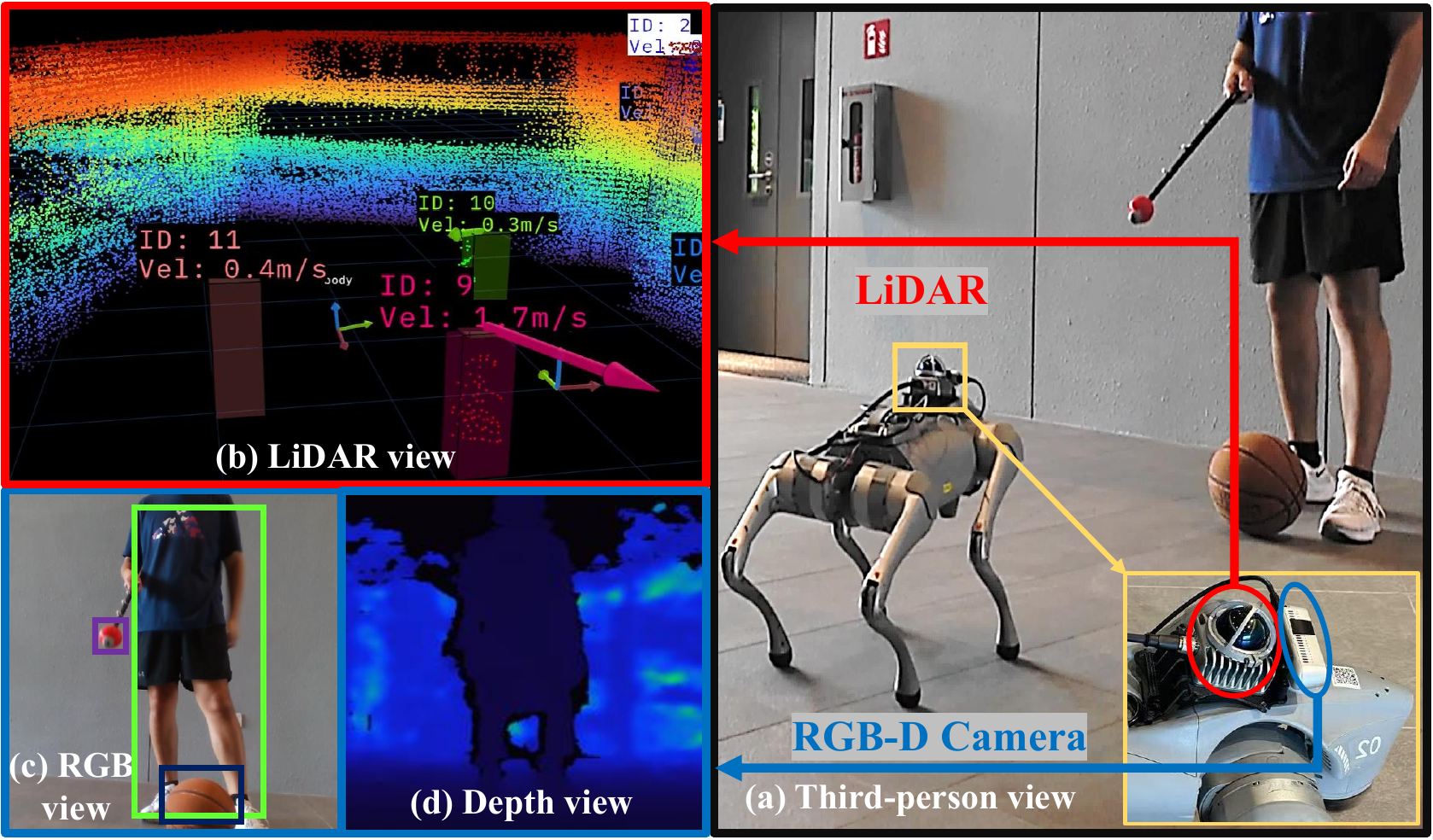}
    \caption{Overview of APREBot’s active perception system. As illustrated in (a), the quadruped robot integrates LiDAR and RGB-D camera to realize active perception: as obstacles approach, the robot combines omnidirectional scanning with active focusing to enable threat-aware avoidance. The LiDAR-scanned point cloud in (b) provides omnidirectional perception, the RGB camera view in (c) enables high-resolution segmentation, and the depth camera view in (d) supports accurate 3D localization of the obstacle.}
    \label{Fig: teaser}
\vspace{-5mm}
\end{figure}

We propose \textbf{APREBot} (Active Perception System for Reflexive Evasion Robot), a novel framework that integrates reflexive evasion with active hierarchical perception to address dynamic obstacle avoidance under constrained reaction time. APREBot employs LiDAR-based omnidirectional scanning for global monitoring to ensure that obstacles approaching from arbitrary directions, including blind spots, can be reliably detected and tracked in real time.  Once a candidate obstacle is identified, camera-driven active focusing is engaged to acquire high-resolution appearance details to distinguish high-risk obstacles (Fig.~\ref{Fig: teaser}). A principled threat level formulation fuses these perception streams to assess obstacle severity and drive adaptive evasion. This perception-driven design builds on REBot~\cite{xu2025rebot}, which demonstrated low-latency reflexive avoidance with privileged obstacle states, and extends it to real-world deployment without reliance on external devices.

We validate APREBot through extensive sim2real experiments on a quadruped platform, covering diverse dynamic obstacle conditions, including obstacles approaching from different directions, with varying speeds, and across multiple object types. The evaluation compares APREBot against state-of-the-art perception baselines, highlighting the advantages of active hierarchical perception and threat-aware avoidance. Results demonstrate that APREBot consistently improves safety margins and reaction efficiency, while preserving the agility required for instantaneous evasive maneuvers.

In summary, the contributions of this paper are as follows:

\begin{itemize}
    \item We propose APREBot, an active hierarchical perception framework that enables quadrupeds to perceive dynamic obstacles, evaluate their threat levels, and perform reflexive evasion without privileged information.  

    \item We develop threat-aware perception mechanisms that combine LiDAR omnidirectional scanning, camera-based active tracking, short-horizon prediction, and a principled threat level for adaptive avoidance.  

    \item We validate APREBot through extensive sim2real experiments, showing consistent improvements over representative perception baselines in safety, efficiency, and robustness.  
\end{itemize}

\sectionreducemargin{Related Works} \label{Sec: related}

Single-sensor systems are widely explored for perception. Raw LiDAR methods, such as Omni-Perception~\cite{wang2025omni}, provide 360° geometric coverage but point clouds are inherently sparse, noisy, and distorted by motion, which hinders robust detection and tracking. Learning-based LiDAR detectors have advanced in autonomous driving~\cite{lu2024fapp,fan2025flying}, yet suffer from domain gaps and poor cross-sensor generalization across platforms such as Avia, Velodyne, and MID-360. Vision-based approaches, including ABS~\cite{he2024agile} and dynamic interaction tasks such as badminton robots~\cite{ma2025learning}, provide semantically rich and smooth signals that support precise tracking. However, when applied to dynamic obstacle avoidance (DOA), LiDAR-only methods struggle with robust real-time detection and tracking~\cite{schmid2023dynablox}, while vision-only methods remain constrained by their limited field of view and vulnerability to motion blur and occlusion.

Multi-sensor fusion has been investigated to overcome single-sensor limitations. Event-camera and frame-camera pipelines combine microsecond-level temporal resolution with semantic cues, improving robustness under rapid motion~\cite{zhudynamic}, but remain sensitive to illumination with limited semantic understanding. LiDAR-camera fusion approaches~\cite{koide2023general,yuan2021pixel}, including IRS-based alignment~\cite{chen2025irs,11127719} and tightly coupled systems such as R2LIVE~\cite{lin2102r2live} and R3LIVE~\cite{lin2109r3live}, enhance accuracy and robustness through joint optimization at the expense of computational efficiency. While recent works like LV-DOT~\cite{xu2025lv} achieved real-time LiDAR-RGB fusion, existing methods still exhibit limitations in matching strategies, motion pattern filtering, and underutilize RGB-based 2D tracking robustness.
\sectionreducemargin{Preliminary} \label{Sec: prelim}

\textbf{Problem formulation.}
We study dynamic obstacle avoidance (DOA) for quadruped robots under a limited reaction time budget $\Delta t_{\text{react}}$.
Each obstacle is modeled as a rigid body with radius $r^O$ and states $\{p^O_t, v^O_t\}$ in the world frame $\mathcal{W}$.
The robot (Unitree Go2) is a high-dimensional articulated system with base states $\{p^R_t, v^R_t, \omega^R_t, \theta^R_t\}$ and leg states $\{q^R_t, \dot q^R_t\}$ (12-DoF).
Unless otherwise specified, all positions and velocities are expressed in the world frame $\mathcal{W}$; orientation and joint states are expressed in the body-fixed frame $\mathcal{B}$. The complete summary of the notation is provided in Tab.~\ref{tab:notation}.

\textbf{Active and hierarchical perception.}
APREBot relies solely on onboard sensing for obstacle perception.
The LiDAR provides omnidirectional point clouds, which are segmented and tracked to obtain coarse obstacle estimates $\{\hat p^O_{t,L}, \hat v^O_{t,L}, \hat r^O_L, id\}$.
Threat levels are computed for all detected obstacles, and the one with the largest LiDAR threat is used to adjust the robot’s direction so that the corresponding obstacle enters the camera field of view.
Once visible, the RGB-D camera provides refined estimates $\{\hat p^O_{t,C}, \hat v^O_{t,C}, \hat r^O_C, id\}$.
This hierarchical perception pipeline exploits complementary sensing roles: LiDAR ensures $360^\circ$ coverage with coarse estimates, while the camera contributes high-precision states for the most threatening obstacle.

\textbf{Threat level.}
For a source $s \in \{L,C\}$ (LiDAR or Camera), we define a unified threat metric as in Eq.~\ref{eq:threat}:
\begin{equation}
\label{eq:threat}
\mathcal{T}^{s}_t
= \alpha \frac{\max\!\big(0,\ v^{s}_{\text{rel},t}\cdot \hat d^{s}_t\big)}{\|\Delta p^{s}_t\|+\epsilon}
+ (1-\alpha)\left(\frac{r_{\text{safe}}}{\|\Delta p^{s}_t\|+\epsilon}\right)^{\gamma},
\end{equation}
where $\Delta p^{s}_t=\hat p^O_{t,s}-p^R_t$ is the relative position, 
$v^{s}_{\text{rel},t}=\hat v^O_{t,s}-v^R_t$ is the relative velocity, 
and $\hat d^{s}_t=\Delta p^{s}_t/\|\Delta p^{s}_t\|$ is the unit approach direction.
The first term captures approach speed normalized by distance, while the second provides a proximity prior that yields nonzero threat for slow but nearby obstacles. 
Parameters $\alpha\in[0,1]$, $\gamma\geq 1$, and $r_{\text{safe}}$ control the balance between these two effects, and $\epsilon>0$ avoids division by zero.
LiDAR-based $\mathcal{T}^L_t$ is always available; when the camera is engaged, $\mathcal{T}^C_t$ is used for the refined target.
In multi-obstacle scenarios, the maximum threat across obstacles is selected to indicate the most urgent direction. 
Compared with the classical time-to-collision (TTC), Eq.~\eqref{eq:threat} offers greater robustness by remaining well-defined for slow or static yet proximate obstacles, while enabling a principled trade-off between distance- and velocity-based risk components.

\begin{table}[t]
\centering
\begin{threeparttable}
\caption{Notation Summary} \label{tab:notation}
\begin{tabular}{ll}
\toprule
Symbol & Description \\
\midrule
$p^R \in \mathbb{R}^3$ & Robot base position (in $\mathcal{W}$) \\
$v^R \in \mathbb{R}^3$ & Robot base linear velocity (in $\mathcal{W}$) \\
$\omega^R \in \mathbb{R}^3$ & Robot base angular velocity (in $\mathcal{W}$) \\
$\theta^R \in \mathbb{R}^3$ & Robot base orientation (Euler angles, in $\mathcal{B}$) \\
$q^R, \dot q^R \in \mathbb{R}^{12}$ & Joint positions and velocities \\
$p^O, v^O \in \mathbb{R}^3$ & Obstacle position and velocity (in $\mathcal{W}$) \\
$r^O \in \mathbb{R}$ & Obstacle radius \\
$\hat p^O_s, \hat v^O_s \in \mathbb{R}^3$ & Estimated obstacle states from $s\!\in\!\{L,C\}$ \\
$\hat r^O_s \in \mathbb{R}$ & Estimated obstacle radius from $s\!\in\!\{L,C\}$ \\
$id \in \mathbb{N}$ & Obstacle track identity ($ L\&C $) \\
$\Delta p^s \in \mathbb{R}^3$ & Relative position, $\hat p^O_s - p^R$ \\
$v^s_{\text{rel}} \in \mathbb{R}^3$ & Relative velocity, $\hat v^O_s - v^R$ \\
$\hat d^s \in \mathbb{R}^3$ & Unit approach vector, $\Delta p^s/\|\Delta p^s\|$ \\
$\mathcal{T}^L, \mathcal{T}^C \in \mathbb{R}_{\ge 0}$ & LiDAR / Camera threat levels \\
$r_{\text{safe}}, \alpha, \gamma, \epsilon \in \mathbb{R}_{>0}$ & Proximity radius, weight, exponent, constant \\
\bottomrule
\end{tabular}
\begin{tablenotes}[flushleft]
\begin{small}
\item World/body frames $\mathcal{W}/\mathcal{B}$. Time index $t$ omitted when clear.
\end{small}
\end{tablenotes}
\end{threeparttable}
\vspace{-5mm}
\end{table}

\sectionreducemargin{Method} \label{Sec: method}
In this section, we present APREBot, a hierarchical perception and threat-aware avoidance framework for dynamic obstacle avoidance. We outline the system structure (Sec.~\ref{Subsec: System structure}) with three stages—omnidirectional scanning, active tracking, and threat-aware avoidance—coordinated by a unified threat metric; describe hierarchical perception (Sec.~\ref{Subsec: Perception}); and the threat-aware avoidance policy (Sec.~\ref{Subsec: Threat-aware avoidance policy}).

\begin{figure*}[t]
    \centering
    \includegraphics[width=0.97\textwidth]{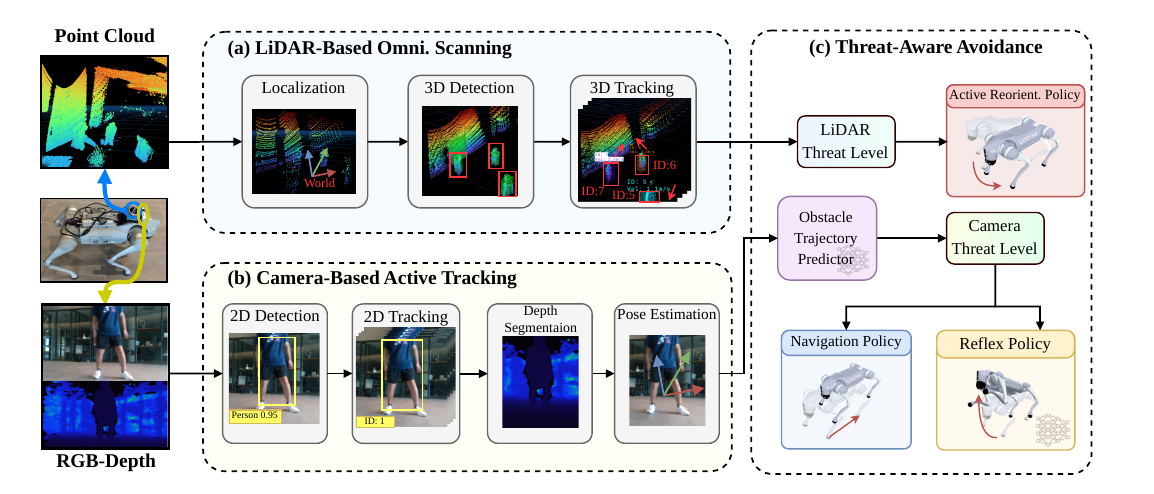}
    \caption{Overview of the APREBot framework consisting of three stages. \textbf{(a) Omnidirectional Scanning Stage:} LiDAR-based perception operates on point cloud to perform localization and provides surrounding obstacle states through 3D detection and tracking. \textbf{(b) Active Tracking Stage:} The robot focuses on the most threatening obstacle, where the onboard RGB-D camera performs high-resolution 2D detection and tracking for segmentation, and uses depth-based refinement for accurate 3D state estimation. \textbf{(c) Threat-Aware Avoidance Stage:} Based on the evaluated threat level, the robot executes appropriate strategies, including active reorientation, navigation-based avoidance, and reflexive evasion.}
    \label{Fig: pipeline}
\vspace{-5mm}
\end{figure*}

\subsectionreducemargin{System structure} \label{Subsec: System structure}

APREBot executes a three-stage perception--control cycle under the reaction-time budget: Omnidirectional Scanning Stage, Active Tracking Stage, and Threat-Aware Avoidance Stage (Fig.~\ref{Fig: pipeline}).

\textbf{(a) Omnidirectional scanning stage.}
The onboard LiDAR continuously segments and tracks obstacles, yielding coarse estimates 
$\{\hat p^O_{t,L}, \hat v^O_{t,L}, \hat r^O_L, id\}$ together with obstacle threats $\mathcal{T}^L_t$. 
The target is identified as the obstacle with the largest $\mathcal{T}^L_t$, which establishes a global priority while preserving omnidirectional situational awareness. 
Target identity is switched only if another obstacle keeps a higher threat for several consecutive steps, avoiding frequent changes in cluttered scenes; the approach direction $\hat d^L_t=\Delta p^L_t/\|\Delta p^L_t\|$ provides an orientation signal for subsequent acquisition.

\textbf{(b) Active tracking stage.}
The robot adjusts its base orientation $\theta^R_t$ (and angular velocity $\omega^R_t$) so that the active target enters the RGB-D camera field of view.
Once visible, the camera provides refined segmentation of different objects (e.g., humans and balls) and estimates $\{\hat p^O_{t,C}, \hat v^O_{t,C}, \hat r^O_C, id\}$, with the corresponding threat $\mathcal{T}^C_t$ replacing the coarse LiDAR quantities as primary inputs to decision-making.
LiDAR continues to furnish panoramic coverage; if a newly detected obstacle exhibits a consistently larger $\mathcal{T}^L_t$ than the current target, the perception focus is reassigned.

\textbf{(c) Threat-aware avoidance stage.}
The entire avoidance process is driven by threat level. LiDAR first identifies the obstacle with the highest threat and triggers active reorientation to bring it into the camera field of view. The camera then refines the threat-level assessment: lower levels lead to navigation-based retreat, while higher levels trigger reflexive evasions such as jumping or crouching.

This three-stage design preserves omnidirectional awareness, allocates selective refinement to the most critical target, and yields avoidance behaviors that adapt to Eq.~\eqref{eq:threat} under the timing constraint $\Delta t_{\text{react}}$.

\subsectionreducemargin{Hierarchical perception} \label{Subsec: Perception}

APREBot employs a hierarchical perception system that combines the complementary strengths of LiDAR and cameras.  LiDAR provides omnidirectional coverage, enabling reliable detection and tracking of obstacles approaching from any direction.  The camera is then actively oriented toward selected targets to perform recognition and depth-based localization, allowing the system to distinguish obstacle types and refine their positions..

\textbf{LiDAR-based omnidirectional scanning.} Our LiDAR module follows a geometry-driven pipeline, avoiding reliance on domain-specific training data and thus remaining robust across heterogeneous LiDARs and unstructured environments. As shown in Alg.~\ref{alg:lidar_tracking}, the raw point cloud  $\mathcal{P}_t$ is first registered to the world frame using FAST-LIO, then downsampled and cropped by region-of-interest filters to reduce noise and computational load. Then, DBSCAN is applied to extract obstacle clusters, each represented by a 3D bounding box.

To maintain temporal consistency, bounding boxes across frames are associated through a Hungarian matcher with increased weight on horizontal displacement, reflecting the dominant motion pattern of approaching obstacles. Track identities are updated through a birth–death mechanism, while a motion-consistency filter suppresses spurious tracks that fail position or velocity stability tests. Finally, each valid track is smoothed with a 3D Kalman filter, yielding obstacle states $(\hat p^O_{L}, \hat v^O_{L}, \hat r^O_{L}, id)$ that are maintained online for subsequent threat evaluation and avoidance control.

\newcommand{\INPUT}{\item[\textbf{Input:}]}
\newcommand{\OUTPUT}{\item[\textbf{Output:}]}

\begin{algorithm}[t]
\caption{LiDAR Perception}
\label{alg:lidar_tracking}
\begin{algorithmic}[1]
\INPUT LiDAR point cloud $\mathcal{P}_t$
\OUTPUT Estimated obstacle states $\{\hat p^O_L, \hat v^O_L, \hat r^O_L, id\}$

\STATE $\mathcal{P}_t^{world} \leftarrow \text{FastLIO}(\mathcal{P}_t)$
\STATE $\mathcal{P}_t^{roi} \leftarrow \text{Filtering}(\mathcal{P}_t^{world})$
\STATE $\mathcal{C}_t \leftarrow \text{DBSCAN}(\mathcal{P}_t^{roi})$
\STATE $\mathcal{B}_t^{3D} \leftarrow \text{ComputeBBox}(\mathcal{C}_t)$

\STATE \textbf{/* Hungarian Matching */}
\FOR{$i,j$ in $\mathcal{B}_{t-1}^{3D} \times \mathcal{B}_t^{3D}$}
    \STATE $C_{ij} \leftarrow 2\|\tilde p_{i,xy}^{t-1} - \tilde p_{j,xy}^{t}\|_2 
        + \|\tilde p_{i,z}^{t-1} - \tilde p_{j,z}^{t}\| - \text{IoU}_{ij}$
\ENDFOR
\STATE $\mathcal{M}_t \leftarrow \text{Hungarian}(C)$
\STATE $\mathcal{B}_t^{tracked} \leftarrow \text{AssignIDs}(\mathcal{B}_t^{3D}, \mathcal{M}_t)$

\STATE \textbf{/* Motion Pattern Filtering */}
\FOR{track $b_i \in \mathcal{B}_t^{tracked}$}
    \STATE $\Sigma_p \leftarrow \text{Cov}(\{\tilde p_i^{t-k:t}\})$
    \STATE $\sigma_v \leftarrow \text{Std}(\{\|v_i^{t-k:t}\|\})$
    \IF{$\mathrm{tr}(\Sigma_p) > \epsilon_p$ OR $\sigma_v > \epsilon_v$}
        \STATE Keep track $b_i$
    \ENDIF
\ENDFOR

\STATE \textbf{/* Kalman Smoothing */}
\STATE $\{\hat p^O_L, \hat v^O_L, \hat r^O_L, id\} 
        \leftarrow \text{KalmanFilter3D}(\mathcal{B}_t^{filtered})$

\RETURN $\{\hat p^O_L, \hat v^O_L, \hat r^O_L, id\}$
\end{algorithmic}
\end{algorithm}

\begin{algorithm}[t]
\caption{RGB-Depth Perception}
\label{alg:rgb_pose_estimation}
\begin{algorithmic}[1]
\INPUT RGB image $\mathcal{I}_t^{rgb}$, Depth image $\mathcal{I}_t^{depth}$, Camera params $\mathbf{K}, [\mathbf{R}|\mathbf{t}]$
\OUTPUT Estimated obstacle states $\{\hat p^O_C, \hat v^O_C, \hat r^O_C, id\}$

\STATE \textbf{/* 2D Detection and Tracking */}
\STATE $\mathcal{D}_t^{2D} \leftarrow \text{YOLO}(\mathcal{I}_t^{rgb})$
\STATE $\mathcal{D}_t^{tracked} \leftarrow \text{ByteTrack}(\mathcal{D}_t^{2D})$

\STATE \textbf{/* BFS Instance Segmentation */}
\FOR{detection $d_i \in \mathcal{D}_t^{tracked}$}
    \STATE $c_i \leftarrow \text{BBoxCenter}(d_i)$
    \STATE $\mathcal{M}_i \leftarrow \text{BFS}(c_i, \mathcal{I}_t^{depth}, \tau_{depth})$
    \STATE \textbf{/* 3D Pose Estimation */}
    \STATE $\mathcal{P}_i^{cam} \leftarrow \{\mathbf{K}^{-1}[u \cdot z, v \cdot z, z]^T : (u,v) \in \mathcal{M}_i, z = \mathcal{I}_t^{depth}(u,v)\}$
    \STATE $\tilde p_i^{cam} \leftarrow \text{Centroid}(\mathcal{P}_i^{cam})$
    \STATE $\hat p^O_{C,i} \leftarrow \mathbf{R}\,\tilde p_i^{cam} + \mathbf{t}$
    \STATE $\hat r^O_{C,i} \leftarrow \text{EstimateRadius}(\mathcal{P}_i^{cam})$
    
    \STATE \textbf{/* Velocity Estimation */}
    \STATE $\hat v^O_{C,i} \leftarrow \text{Diff}(\hat p^O_{C,i})$
\ENDFOR

\RETURN $\{\hat p^O_C, \hat v^O_C, \hat r^O_C, id\}$
\end{algorithmic}
\end{algorithm}

\textbf{RGB-D-based active focusing.}
When an active target enters the camera’s field of view, an RGB-D pipeline is activated to provide precise and metrically consistent obstacle states. As summarized in Alg.~\ref{alg:rgb_pose_estimation}, object candidates are detected on RGB frames by YOLO and temporally associated with ByteTrack to yield stable 2D tracks. For 3D localization, each track is refined by a lightweight center-seeded breadth-first search on the depth image, expanding a mask of depth-consistent pixels. The resulting region is back-projected through camera intrinsics and extrinsics to obtain a 3D point cloud, from which both the centroid and an effective radius are estimated.

Obstacle velocities are then computed via temporal differencing, completing the state estimate $(\hat p^O_{C}, \hat v^O_{C}, \hat r^O_{C}, id)$. This design enables the camera pathway to complement LiDAR by providing high-resolution 3D localization and geometric size estimates whenever the obstacle is in view, while remaining efficient enough for real-time onboard deployment.

\textbf{Short-horizon prediction.}
To anticipate obstacle motion, we employ an LSTM predictor trained on simulated trajectories. The network takes as input a short history of obstacle positions and velocities \{${p^O_{(t-k:t)}, v^O_{(t-k:t)}}$\} over several frames, and outputs one-step forecasts \{${\hat p^O_{t+1}, \hat v^O_{t+1}}$\}. These predictions align with the perception outputs and provide forward-looking estimates of obstacle states under the constrained reaction horizon.

\subsectionreducemargin{Threat-aware avoidance policy} \label{Subsec: Threat-aware avoidance policy}

The controller uses the robot state $s^R_t$ and the perceived obstacle state $\{\hat p^O_t,\hat v^O_t,\hat r^O,id\}$ together with the unified threat $\mathcal{T}_t$.
A monotone schedule $\beta_t=g(\mathcal{T}_t)\in[0,1]$ continuously blends a conservative navigation controller with an aggressive reflex policy, while a lightweight turning module keeps the camera aligned with the most threatening direction.
Commands are finally blended as $a_t=(1-\beta_t)\,a^{\mathrm{nav}}_t+\beta_t\,a^{\mathrm{reflex}}_t$.

\textbf{Active reorientation policy.} Active reorientation is tied to the threat gap between LiDAR’s omnidirectional estimate and the current fused threat, so it only acts when LiDAR reveals a more urgent direction.
Let $\phi_t=\operatorname{atan2}((\hat p^O_t-p^R_t)_y,(\hat p^O_t-p^R_t)_x)$ be the target direction and $\psi^R_t$ the yaw in $\theta^R_t$.
We track the heading with a bounded yaw–rate command scaled by the gap:
\begin{equation}
    \dot\psi^{\mathrm{ref}}_t
=\operatorname{sat}\!\Big(k_\psi\,\big[\mathcal{T}^L_t-\mathcal{T}_t\big]_+\,(\phi_t-\psi^R_t)\Big),
\end{equation}
where $[x]_+=\max(x,0)$ and $\operatorname{sat}(\cdot)$ limits excessive spin and $k_\psi>0$ is the yaw–rate gain.
Thus, when the camera’s target already dominates ($\mathcal{T}_t\!\ge\!\mathcal{T}^L_t$), turning naturally fades; if LiDAR sees a higher threat elsewhere, the gap grows and the robot reorients.

\textbf{Navigation-based avoidance policy.} Navigation-based avoidance is a threat-aware reference-tracking behavior that enlarges clearance under moderate risk.
Let $\hat d_t=(p^R_t-\hat p^O_t)/\|\hat p^O_t-p^R_t\|$ be the obstacle-to-robot direction.
We set a planar velocity reference opposite to the approach,
\begin{equation}
    v^{\mathrm{ref}}_{t,xy}
= -\,\Big(k_r^{\min}+(k_r^{\max}-k_r^{\min})\,g(\mathcal{T}^C_t)\Big)\,\hat d_{t,xy},
\end{equation}
and track it in the locomotion stack. Here $k_r^{\min}$ and $k_r^{\max}$ are the lower and upper bounds of the velocity gain.
As $\mathcal{T}^C_t$ increases, the gain grows smoothly, yielding longer, faster steps; LiDAR-driven updates ensure that the commanded velocity adaptively reorients according to the evolving threat direction.

\textbf{Reflexive evasion policy.} When $\mathcal{T}_t$ is high or the remaining reaction time is short, control blends toward a learned reflex policy, adapted from the REBot baseline~\cite{xu2025rebot}, that produces rapid evasive maneuvers (e.g., sidesteps, crouches, jumps)..
Perception is augmented with a short-horizon prediction $\hat p^O_{t+1}$ to compensate sensing/actuation delays, and response intensity scales with $\mathcal{T}_t$ via $g(\mathcal{T}_t)$.
The reflex objective is a compact, threat-aware composite reward
\begin{equation}
    r_t=\lambda_{\text{safe}} r^{\text{safe}}_t
    +\lambda_{\text{dir}}  r^{\text{dir}}_t
    +\lambda_{\text{ene}}  r^{\text{ene}}_t
    +\lambda_{\text{stab}} r^{\text{stab}}_t
    +\lambda_{\text{rec}}  r^{\text{rec}}_t.
\end{equation}

Safety $r^{\text{safe}}_t$ penalizes small predicted clearance to the obstacle, pushing the robot to maintain a safe margin even under short reaction time. Direction $r^{\text{dir}}_t$ discourages velocity components toward the obstacle and encourages net displacement away from the approach direction, reducing circling or oscillatory motion. Energy $r^{\text{ene}}_t$ suppresses unnecessary actuation to keep evasive responses efficient rather than brute-force. Stability $r^{\text{stab}}_t$ maintains dynamic steadiness of the base and reliable support during rapid maneuvers, improving controllability and preventing tumbles. Recovery $r^{\text{rec}}_t$ encourages a quick return to stable posture and low velocities once the threat subsides, enabling smooth handover back to normal operation.

All terms are uniformly scaled by $g(\mathcal{T}_t)$, while the reorientation and navigation-based controllers remain purely reference-tracking and independent of this reward.

\sectionreducemargin{Experiment} \label{Sec: experiment}

We evaluate APREBot on a quadruped platform with fully onboard sensing and control. The study proceeds from experimental setup and tasks descriptions (Sec.~\ref{Subsec: Experiment settings}), through policy training in simulation and deployment on real robot (Sec.~\ref{Subsec: Experiment Implementation}), followed by main results analysis (Sec.~\ref{Subsec: Main Results}), and ablations (Sec.~\ref{Subsec: Ablation Studies}) that isolate critical components.

Our assessment is organized around four research questions: \textbf{Q1}: Does LiDAR-based detection and tracking outperform purely reactive use of raw point-cloud scans?  \textbf{Q2}: Does hierarchical perception---LiDAR for omnidirectional scanning combined with camera-based active focusing---yield superior performance compared to LiDAR-only sensing?  \textbf{Q3}: Does active reorientation toward the most threatening target improve performance, particularly for rear or blind-spot approaches?  \textbf{Q4}: Given active reorientation, does threat-aware scheduling provide advantages over fixed rule--based triggering schemes (e.g., TTC thresholds)?

\begin{figure*}[t]
    \centering
    \includegraphics[width=0.90\textwidth]{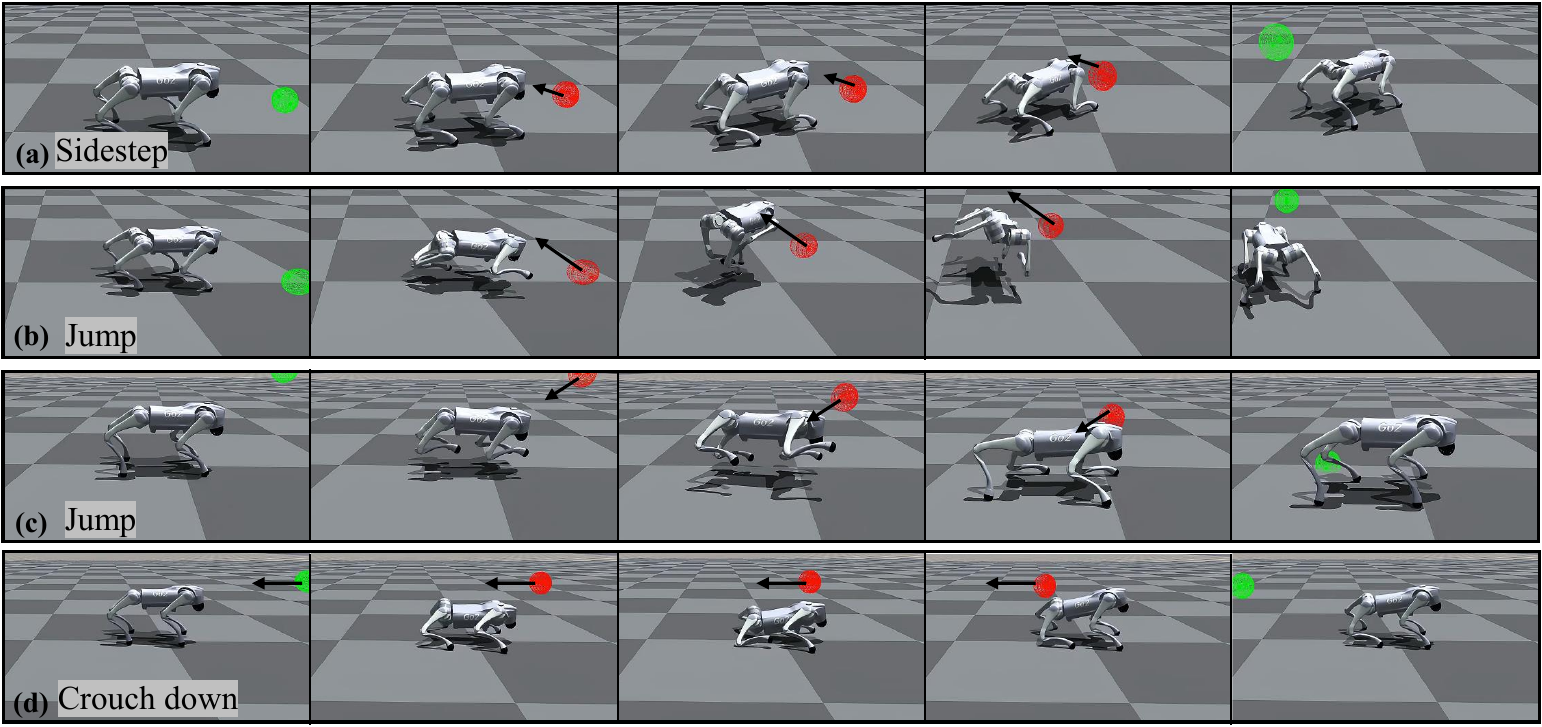}
    \caption{Simulation experiment results of reflexive policy showing different evasion gaits. \textcolor{green}{\CIRCLE} static obstacle; \textcolor{red}{\CIRCLE} dynamic obstacle. (a) Sidestep to avoid the obstacle. (b-c) Jump to avoid the obstacle. (d) Crouch down to avoid the obstacle.}
    \label{Fig: experiments}
\vspace{-2mm}
\end{figure*}

\subsectionreducemargin{Experiment settings} \label{Subsec: Experiment settings}

\textbf{Baselines.} 
We compare against three representative LiDAR-only baselines, covering both geometry-driven and learning-based approaches: 
1) \textit{Raycast}~\cite{pantic2023obstacle}: a method that directly uses raw LiDAR point clouds to generate control commands, without detection or tracking. 
2) \textit{FAPP}~\cite{lu2024fapp}: a method that applies DBSCAN clustering and Kalman filtering for explicit obstacle detection and tracking. 
3) \textit{Omni}~\cite{wang2025omni}: a learning-based method that directly processes spatio-temporal point clouds with the PD-RiskNet perception module, enabling end-to-end omnidirectional avoidance.

\textbf{Tasks.} Obstacles are introduced from diverse directions around the robot, including rear and lateral blind zones outside the forward camera’s field of view. Their motion patterns span both slow, steady approaches and sudden accelerations. The obstacle set includes humans, balls, and sticks. In each trial, initial position, approach direction, and velocity are randomized within predefined ranges to ensure a broad distribution of threat conditions.

\noindent \textbf{Metrics.} We evaluate performance on 4 system-level metrics: 

\begin{itemize}
    \item Avoidance Success Rate (\textit{ASR}). Defined by 
    $\text{ASR}=N_{\text{avoid}}/N_{\text{total}}$, 
    where $N_{\text{avoid}}$ is the number of successful trials (no collision and stable recovery) out of $N_{\text{total}}$ total trials. 

    \item Minimum clearance $d_{\min}$. Computed as 
    $d_{\min}=\min_t(\|p_o(t)-p_r(t)\|-R_s)$, 
    where $p_o(t)$ and $p_r(t)$ denote obstacle and robot base positions, and $R_s$ is the combined safety radius. 

    \item Normalized Trigger Lead (\textit{TNL}). Defined by 
    $\text{TNL}=(t_{\text{closest}}-t_{\text{trig}})/\text{TTC}(t_{\text{trig}})$, 
    where $t_{\text{trig}}$ is the trigger time, $t_{\text{closest}}$ the time of minimum distance, and $\text{TTC}(t_{\text{trig}})$ the estimated time-to-contact at trigger. 
    We compute $\text{TTC}(t)$ under a constant-velocity assumption:
    \begin{equation}
           \text{TTC}(t) = \frac{\|\Delta p^s_t\|}{\max\!\big(0,\, v_{\text{rel}}^s(t)\cdot \hat d^s(t)\big) + \varepsilon}, 
    \end{equation}
    where $\Delta p^s_t = p^O_t - p^R_t$ is the obstacle--robot displacement in the spatial frame, 
    $\hat d^s(t)=\Delta p^s_t/\|\Delta p^s_t\|$ its unit direction, and $v_{\text{rel}}^s(t)$ the relative velocity. 
    A small $\varepsilon$ avoids division by zero. 

    \item Energy $E$. The mechanical work between trigger and recovery: 
    $E=\int_{t_{\text{trig}}}^{t_{\text{rec}}}|\tau(t)^\top \dot q(t)|dt \approx \sum_k |\tau_k^\top \dot q_k|\Delta t$, 
    where $\tau$ and $\dot q$ are joint torques and velocities. 
\end{itemize}

\begin{figure*}[t]
    \centering
    \includegraphics[width=0.97\textwidth]{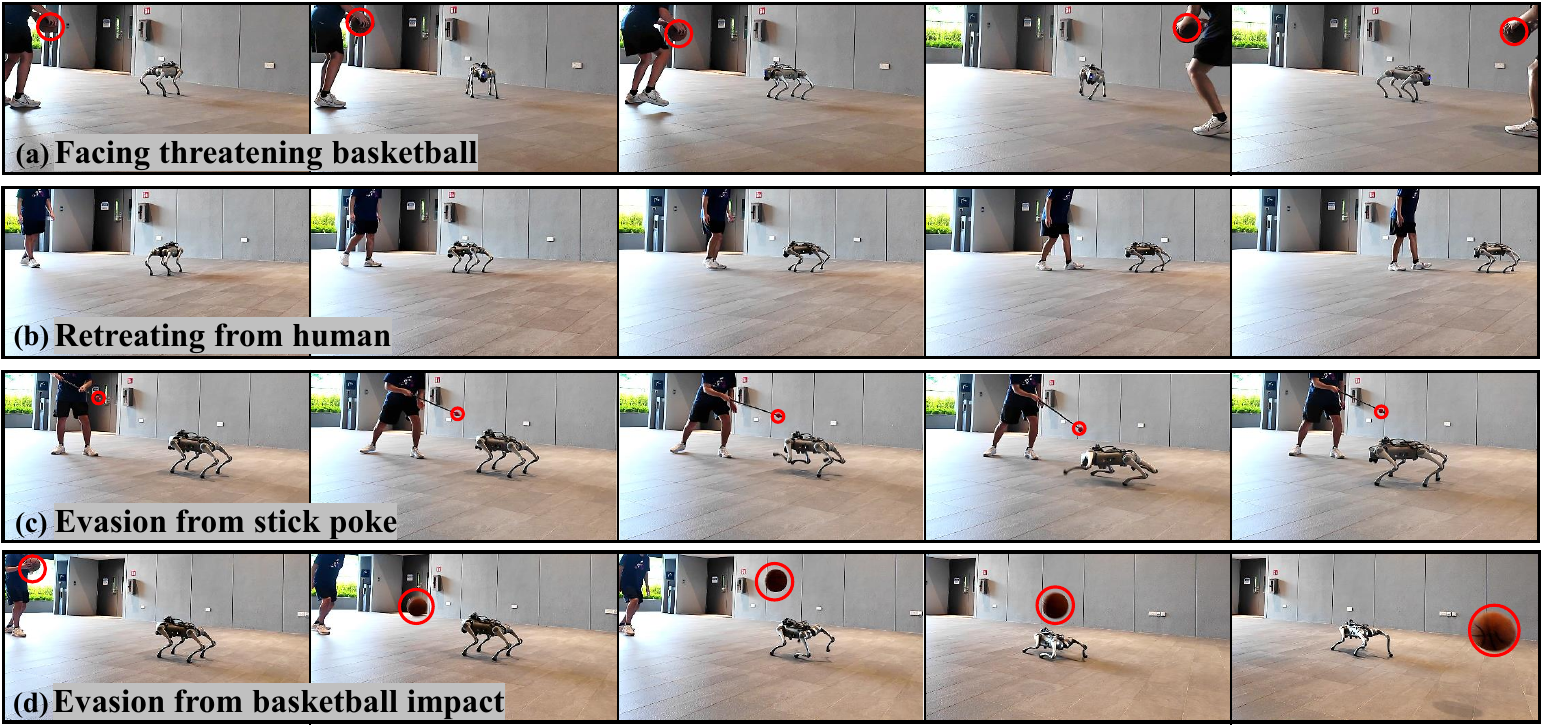}
    \caption{Real-robot experiments demonstrating threat-aware decision making of the robot. (a) The robot identifies the basketball in the human’s hand as a threat and continuously reorients to face it. (b) The robot retreats slowly as the human approaches. (c) When poked rapidly with a stick, the robot performs reflexive evasion. (d) When hit with a basketball, the robot executes reflexive evasion.}
    \label{Fig: real robot}
\vspace{-5mm}
\end{figure*}

\subsectionreducemargin{Experiment Implementation}\label{Subsec: Experiment Implementation}

\textbf{Policy training in simulation}.
We train the policy with PPO in Isaac Gym with 4,096 parallel environments using a Unitree Go2 quadruped model. Dynamic obstacles are modeled as spheres with randomized positions, velocities, and approach directions, ensuring diverse encounter conditions. 
Domain randomization is applied to robot dynamics (mass, friction, joint damping) and sensor noise, improving robustness for sim2real transfer. Policies converge stably within 6,000 iterations before deployment to the real robot.

The results shows the robot demonstrates robust evasive maneuvers under diverse obstacle approaches (see Fig.~\ref{Fig: experiments}). Depending on the incoming direction and remaining reaction time, the quadruped adapts with distinct strategies: lateral sidesteps(Fig.~\ref{Fig: experiments}(a)), sudden jumps(Fig.~\ref{Fig: experiments}(b)(c)), or crouching down(Fig.~\ref{Fig: experiments}(d) to avoid obstacles.

\textbf{Real-robot deployment}.
Experiments are conducted on a Unitree Go2 quadruped equipped with onboard sensing and computation. The Livox MID-360 LiDAR provides $360^\circ$ range data for obstacle detection and tracking, while the Intel RealSense D435 RGB-D camera offers high-resolution obstacle details. An onboard trajectory predictor estimates short-horizon obstacle trajectories, and all modules run in real time on the NVIDIA Jetson Orin without external sensing or offboard computation. Policies trained in simulation are deployed to the real robot directly. 

\begin{figure}[t]
    \centering
    \includegraphics[width=0.97\columnwidth]{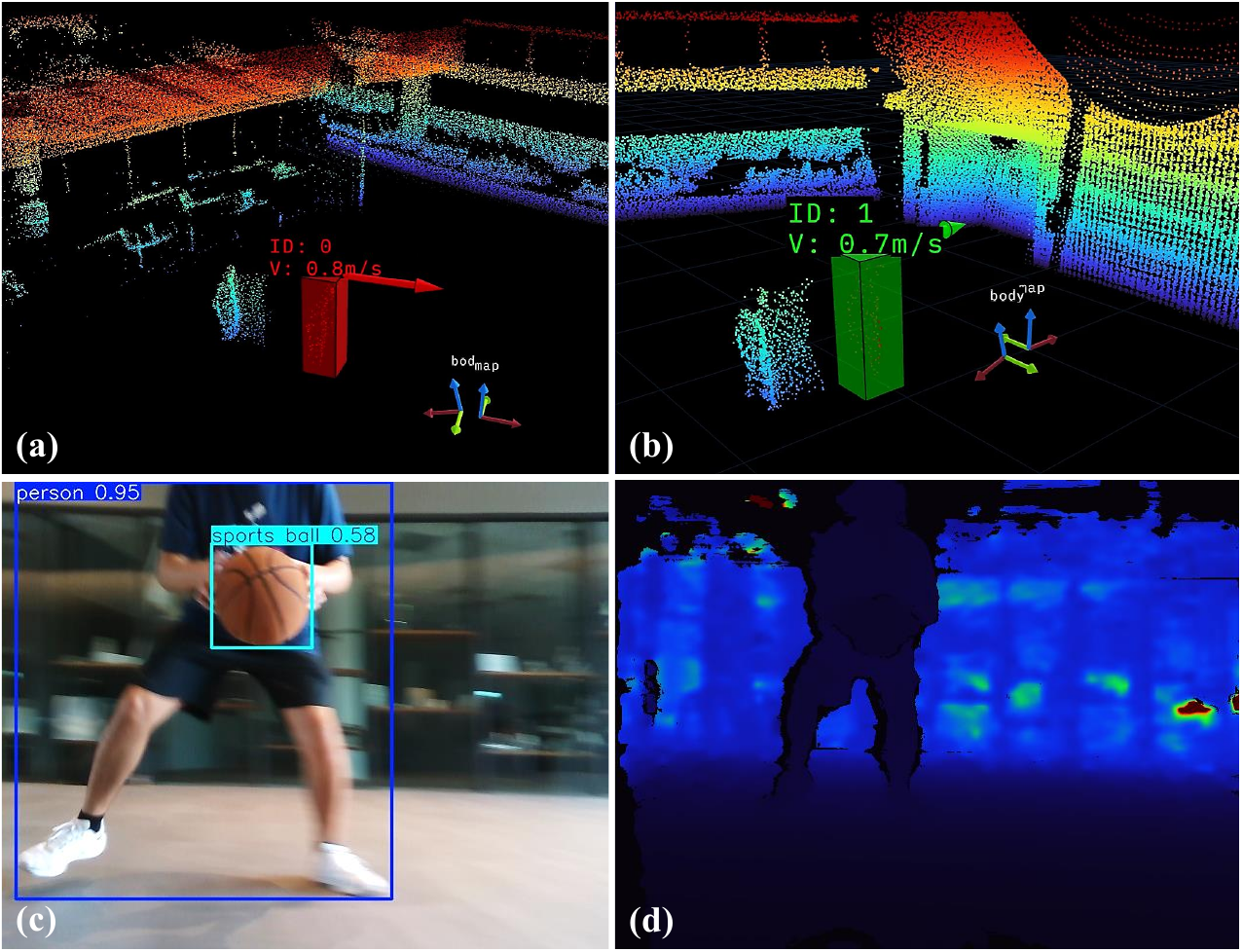}
    \caption{Perception performance of APREBot. (a–b)  LiDAR-based clustering and tracking of surrounding obstacles. (c) Camera RGB image with segmentation of a person and a basketball. (d) Camera depth image used for 3D position estimation.}
    \label{Fig: results2}
\vspace{-3mm}
\end{figure}

Fig.~\ref{Fig: results2} illustrates APREBot’s perception performance. LiDAR enables omnidirectional clustering and tracking of moving obstacles(Fig.~\ref{Fig: results2}(a)(b)), while the RGB-D camera segments objects such as humans and balls(Fig.~\ref{Fig: results2}(c)), and uses depth estimation to recover their 3D positions for precise localization(Fig.~\ref{Fig: results2}(d)).
Fig.~\ref{Fig: real robot} shows representative trials on the quadruped platform. After LiDAR scanning, the robot actively reorients toward threatening obstacles(Fig.~\ref{Fig: real robot}(a)). Under slow approaches it performs navigation-based retreat(Fig.~\ref{Fig: real robot}(b)), whereas fast-approaching threats trigger reflexive evasions(Fig.~\ref{Fig: real robot}(c)(d)), each followed by smooth recovery to its normal state.

\begin{table*}[t]
\centering
\setlength{\tabcolsep}{14pt}
\begin{threeparttable}
\caption{Real-Robot Experiment Results}
\label{tab:main_results}
\begin{tabular}{lcccc}
\toprule
Method & ASR~[$\%$] $\uparrow$ & $d_{\min}$~[m] $\uparrow$ & TNL $\uparrow$ & $E$~[J] $\downarrow$ \\
\midrule
Raycast      & $34.8 \pm 10.2$  & $0.095 \pm 0.030$  & $0.09 \pm 0.05$  & $1250 \pm 220$ \\
FAPP     & $52.1 \pm 9.8$   & $0.140 \pm 0.030$  & $0.15 \pm 0.07$  & $890  \pm 180$ \\
Omni     & $66.9 \pm 8.9$   & $0.135 \pm 0.032$  & $0.22 \pm 0.06$  & $930  \pm 150$ \\
APREBot (ours)               & $78.7 \pm 8.1$   & $0.186 \pm 0.035$  & $0.30 \pm 0.08$  & $640  \pm 140$ \\
\bottomrule
\end{tabular}
\begin{tablenotes}[flushleft]
\begin{small}
 \item ASR is computed over 50 trials, while $d_{\min}$, TNL, and $E$ are averaged over successful avoidance trials only.  
Raycast: raw LiDAR point cloud control. FAPP: LiDAR clustering and tracking. 
Omni: LiDAR end-to-end RL. 
APREBot: our proposed method.
\end{small}
\end{tablenotes}
\end{threeparttable}
\vspace{-3mm}
\end{table*}

\subsectionreducemargin{Main results analysis} \label{Subsec: Main Results}

\begin{figure}[t]
    \centering
    \includegraphics[width=0.97\columnwidth]{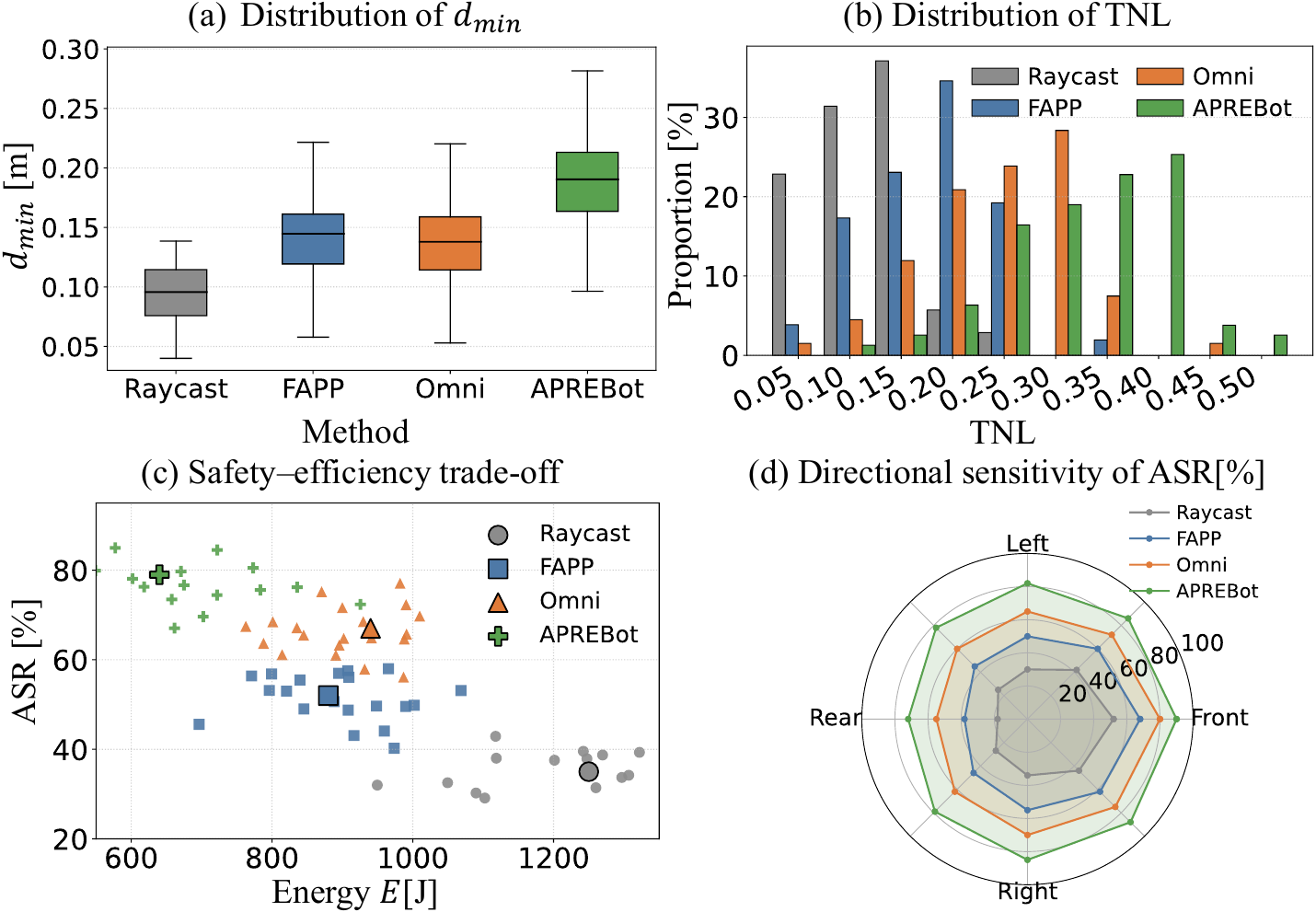}
    \caption{Main experiment results. APREBot consistently outperforms all baselines with larger safety margins, earlier triggers, lower energy cost, and improved blind-side robustness.}
    \label{Fig: results1}
\vspace{-5mm}
\end{figure}

We evaluate three LiDAR-only baselines against APREBot on the real robot using the four metrics: ASR, $d_{\min}$, TNL, and $E$. 
Tab.~\ref{tab:main_results} summarizes the averaged outcomes across randomized trials, and Fig.~\ref{Fig: results1} further illustrates performance differences. Several key findings emerge:

\textbf{LiDAR detection improves over raw point cloud control.}  
Fig.~\ref{Fig: results1}(a) shows that Raycast, which directly maps raw LiDAR scans to reactive actions, yields the smallest $d_{\min}$ and lowest ASR (Tab.~\ref{tab:main_results}). 
By contrast, FAPP introduces explicit obstacle clustering and tracking, significantly increasing ASR and enlarging $d_{\min}$. 
This confirms that structured LiDAR-based detection and tracking provides a more reliable safety margin than raw point cloud control. 

\textbf{Learning-based LiDAR methods improves success, while geometry-based detection is more efficient.}  
As seen in Fig.~\ref{Fig: results1}(a) and Fig.~\ref{Fig: results1}(c), Omni achieves slightly higher ASR and TNL than FAPP, reflecting the strength of end-to-end LiDAR learning methods.  
However, FAPP maintains similar or better $d_{\min}$ and lower energy consumption, indicating that geometry-based detection remains more efficient in terms of energy and stability(Fig.~\ref{Fig: results1}(b)(d)).  
 The results reveal that learning and geometry-based approaches emphasize different priorities, underscoring the value of combining their strengths in future designs.  

\textbf{APREBot outperforms all LiDAR-only baselines.}  
Compared to all baselines, APREBot consistently achieves the highest ASR and largest $d_{\min}$ while reducing energy cost (Fig.~\ref{Fig: results1}(c), Tab.~\ref{tab:main_results}).  
Fig.~\ref{Fig: results1}(b) shows that its TNL distribution shifts to the right, meaning evasive actions are triggered earlier and with larger safety margins.  
Moreover, Fig.~\ref{Fig: results1}(d) reveals that APREBot achieves superior ASR across all approach directions, with especially pronounced gains in rear and blind-side cases, highlighting the benefit of active reorientation and camera-based focusing.  

\subsectionreducemargin{Ablation studies} \label{Subsec: Ablation Studies}

\begin{figure}[t]
    \centering
    \includegraphics[width=0.97\columnwidth]{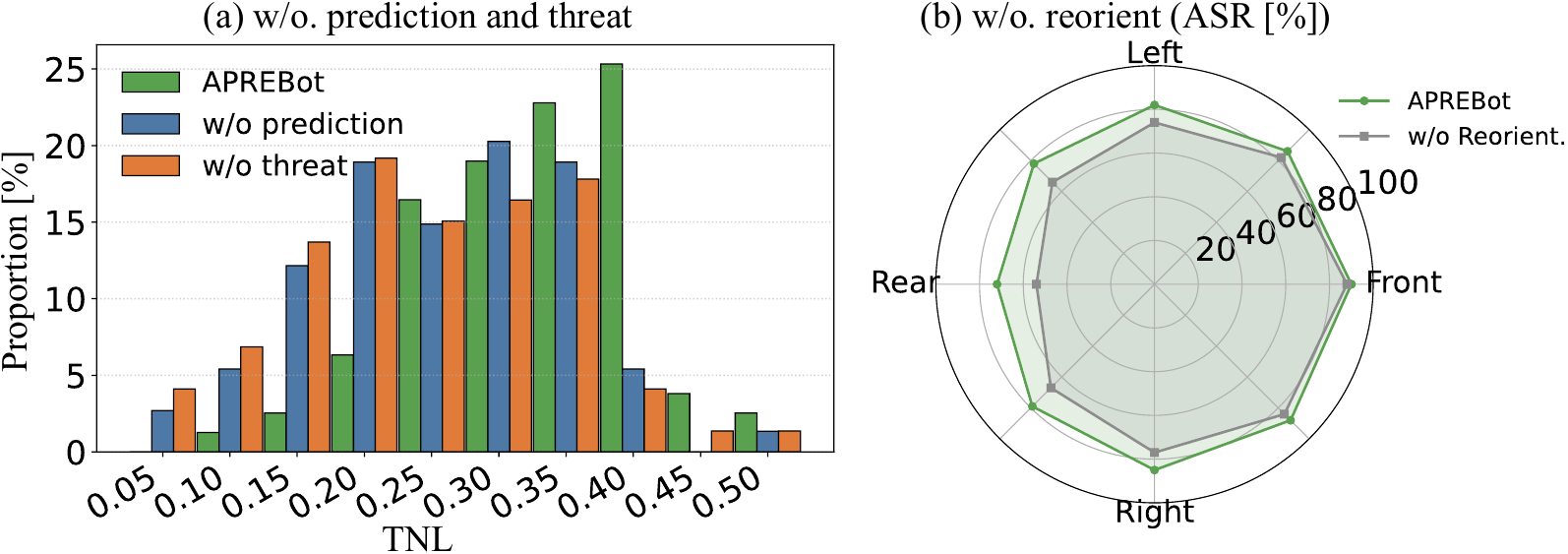}
    \caption{Ablation experiment results.APREBot achieves earlier triggering and larger safety margins compared to variants without prediction or threat, and maintains robust performance in rear and blind-side cases when active reorientation is removed.}
    \label{Fig: ablation}
\vspace{-6mm}
\end{figure}

To assess the role of individual components, we compare APREBot against three ablated variants: removing trajectory prediction (\textit{w/o Prediction}), disabling active reorientation (\textit{w/o Reorient.}), and replacing threat-aware scheduling with a fixed TTC-based rule (\textit{w/o Threat}). Tab.~\ref{tab:ablation_results} reports averaged metrics, while Fig.~\ref{Fig: ablation} illustrates representative differences.

\textbf{Prediction enables earlier responses.}  
Fig.~\ref{Fig: ablation}(a) shows that without prediction, the TNL distribution shifts leftward, indicating later activation. This leads to reduced ASR and smaller safety margins, as the robot occasionally fails against fast-approaching obstacles. With prediction, triggering occurs earlier and evasive maneuvers become more reliable, confirming its role in responsiveness.

\textbf{Active reorientation ensures blind-side robustness.}  
As seen in Fig.~\ref{Fig: ablation}(b), disabling reorientation notably degrades ASR in rear and lateral directions, while frontal performance remains closer to APREBot. By rotating toward the most threatening approach, APREBot brings obstacles into the camera field of view and maintains high success rates across all directions.

\textbf{Threat-aware scheduling improves timing and efficiency.}  
Tab.~\ref{tab:ablation_results} shows that replacing adaptive scheduling with a fixed TTC threshold lowers TNL and raises energy usage. As seen in Fig.~\ref{Fig: ablation}(a), the fixed rule delays activation, so evasive actions occur later and often require more abrupt maneuvers. By contrast, threat-aware scheduling adapts to obstacle velocity and approach direction, enabling earlier and smoother responses.

In summary, prediction supports anticipation, reorientation provides blind-zone coverage, and threat-aware scheduling delivers well-timed and efficient reactions. Their combination gives APREBot the robustness that cannot be achieved by any single component alone.

\begin{table*}[t]
\centering
\caption{Ablation results on the experiments}
\label{tab:ablation_results}
\setlength{\tabcolsep}{14pt}
\begin{threeparttable}
\begin{tabular}{lcccc}
\toprule
Method & ASR~[$\%$] $\uparrow$ & $d_{\min}$~[m] $\uparrow$ & TNL $\uparrow$ & $E$~[J] $\downarrow$\\
\midrule
APREBot (ours)                 & $78.7 \pm 8.1$  & $0.186 \pm 0.035$  & $0.30 \pm 0.08$  & $640 \pm 140$ \\
w/o Prediction           & $74.5 \pm 9.0$  & $0.168 \pm 0.045$  & $0.23 \pm 0.09$  & $700 \pm 150$ \\
w/o Reorient. & $69.8 \pm 9.6$  & $0.148 \pm 0.044$  & $0.25 \pm 0.08$  & $780 \pm 160$ \\
w/o Threat         & $72.9 \pm 9.1$  & $0.160 \pm 0.050$  & $0.22 \pm 0.10$  & $770 \pm 170$ \\
\bottomrule
\end{tabular}
\begin{tablenotes}[flushleft]
\begin{small}
  
\item w/o Prediction: removes short-horizon trajectory prediction.  
w/o Reorient.: disables threat-aware reorientation toward the most threatening target.  
w/o Threat: replaces threat-aware scheduling with a TTC-only triggering rule.
\end{small}
\end{tablenotes}
\end{threeparttable}
\vspace{-5mm}
\end{table*}
\sectionreducemargin{Conclusion} \label{Sec: conclusion}

This work presented APREBot, a quadrupedal system that leverages active perception by integrating omnidirectional LiDAR scanning and camera-based active tracking to enable threat-aware decision making for reflexive evasion. By integrating real-time detection, short-horizon trajectory prediction, and threat-aware scheduling into the control policy, APREBot achieves agile and robust avoidance of dynamic obstacles without reliance on external device. Extensive real-robot experiments demonstrated that APREBot substantially outperforms LiDAR-only baselines and validated its performance across diverse approach directions, including challenging rear and blind-side encounters. These results highlight the importance of active perception and adaptive decision making as foundations for robust quadrupedal reflexive evasion.

Future work will extend the APREBot framework to more general and demanding scenarios, such as evasion during locomotion and navigation in densely cluttered environments. These efforts aim to advance toward fully autonomous quadrupeds capable of safe and agile operation in unstructured and highly dynamic real-world settings.

\bibliographystyle{IEEEtran}
\bibliography{references}

\end{document}